\documentclass[11pt,a4paper]{article}
\usepackage[hyperref]{acl2020}
\aclfinalcopy
\usepackage{times}
\usepackage{latexsym}
\usepackage{amsmath,amssymb,amsfonts}
\usepackage{algorithmic}
\usepackage{multirow}
\usepackage{url}
\usepackage{graphicx}
\graphicspath{{figures/}}

\usepackage{microtype}

\title{Does a Hybrid Neural Network based Feature Selection Model Improve Text Classification?}

\author{Suman Dowlagar \\
  LTRC\\
  IIIT-Hyderabad\\
  \texttt{suman.dowlagar@} \\
  \texttt{research.iiit.ac.in} \\\And
  Radhika Mamidi \\
  LTRC\\
  IIIT-Hyderabad\\
  \texttt{radhika.mamidi@} \\
  \texttt{iiit.ac.in} \\}

\date{}

\begin{document}
\maketitle
\begin{abstract}
Text classification is a fundamental problem in the field of natural language processing. Text classification mainly focuses on giving more importance to all the relevant features that help classify the textual data. Apart from these, the text can have redundant or highly correlated features. These features increase the complexity of the classification algorithm. Thus, many dimensionality reduction methods were proposed with the traditional machine learning classifiers. The use of dimensionality reduction methods with machine learning classifiers has achieved good results. In this paper, we propose a hybrid feature selection method for obtaining relevant features by combining various filter-based feature selection methods and fastText classifier. We then present three ways of implementing a feature selection and neural network pipeline. We observed a reduction in training time when feature selection methods are used along with neural networks. We also observed a slight increase in accuracy on some datasets.
\end{abstract}

\section{Introduction}

Text classification assigns one or more class labels from a predefined set to a document based on its content. Text classification has broad applications in real-world scenarios such as document categorization, news filtering, spam detection, Optical character recognition (OCR), and intent recognition. Giving high weights to relevant features is the objective of text classification.

The field of text classification has gained more interest during the machine learning (ML) era. Many discriminative and generative machine learning classifiers have achieved excellent results in the field of text classification \cite{deng2019feature}. Feature selection and feature extraction methods are often used to reduce high dimensionality \cite{bharti2015hybrid}. Feature extraction generates features from text \cite{agarwal}. Feature selection (FS) selects the most prominent features \cite{4456849}. 

These feature selection and extraction methods are used along with traditional classification algorithms. These methods reduced the curse of dimensionality and increased the classification accuracy \cite{deng2019feature}. 

Recently, deep learning models are used to learn better text representations and to classify the text \cite{minaee2020deep}. Such models include convolutional neural networks (CNN) \cite{kim2014convolutional}, recurrent neural networks (RNN) \cite{hochreiter1997long}, Transformer models \cite{adhikari2019docbert}, and graph convolutional networks (GCN) \cite{yao2019graph}. These NN models capture semantic and syntactic information in local and global word sequences.

Even though the neural networks capture a complex and dense representation of data, the set of words introducing noise in the classifier is still present. Such words add the burden of increased vocabulary, which results in increased textual representation and an increase in the training time of the classifiers \cite{song2011fast}.

Similar to the traditional approaches, we want to understand the effects of using statistical feature selection algorithms beforehand to calculate the features' relevance and then train a fastText text-classification algorithm on those relevant features. Using this feature selection and neural network pipeline, we assume that the complexity of dealing with larger vocabulary decreases. Including feature selection with fastText text-classification helps reduce the classifier's training time and helps the classifier reach better local optima, showing a significant increase in classification accuracy. 

In this work, we analyzed a feature selection and neural network pipeline for text classification. We used a hybrid feature selection method to get a score on relevant features. Using this score, we formulated three methods. The first and second methods deal with modifying the original text by extracting the relevant features. The third method deals with using the feature selection scores and pass it along with the word embeddings. We then observed the effect of feature selection on various neural networks.

The rest of the paper is organized as follows. Section 2 gives a brief review of previous works in the field of feature selection and text classification. Section 3 presents a detailed procedure of the proposed pipeline and presents the experiments and datasets used for our study. Section 4 reports the performance of text classifiers with and without feature selection methods. Section 5 concludes the paper.

\section{Literature Survey}
This section presents a brief description of the neural network (NN) classification algorithms and various feature selection methods.

\subsection{Deep Learning for Text Classification}

Nowadays, various NNs such as CNN, RNN, BERT, and Text GCN achieve state-of-the-art results on text classification. CNN uses 1d convolutions \cite{zhang2015character} and character level convolutions \cite{conneau2016very} to learn the semantic similarity of words or characters, which helps in classifying the text. RNN models such as GRU, LSTM, and BiLSTM \cite{liu2016recurrent} take word to word sequences
to learn a better textual representation of a document that helps in text classification. Attention mechanisms have been introduced in these LSTM models, which increased the representativeness of
the text for better classification \cite{yang2016hierarchical}. Transformer models such as BERT \cite{devlin2018bert} uses the attention mechanism that learns contextual relations between words or sub-words in a text \cite{adhikari2019docbert}. Text GCN \cite{yao2019graph} uses a graph-convolutional network to learn a heterogeneous word document graph on the whole corpus. Text GCN can capture global word co-occurrence information and use graph convolutions to learn a global representation, which helps classify the documents.

\subsection{Feature Selection on Text Data}

The text classification often involves extensive data with thousands of features. Although tens of thousands of words are in a typical text collection, most of them contain little or no information to predict
the text label. These features introduce complexity and increase the training time of an ML classifier. Feature selection is one method for giving high scores to relevant features (Deng et al., 2019). The goal of feature selection is to select
highly-relevant features with minimum redundancy. The relevance of a feature indicates that the feature is always necessary to predict the class label.

There are various text feature selection methods in the literature, each being filter, wrapper, hybrid, and embedded methods. The filter method evaluates the quality of a feature using a scoring function. Some filter methods evaluate the goodness of a term based on how frequently it appears in a text corpus. Document Frequency (DF) \cite{lam1999feature} and Term Frequency - Inverse Document Frequency (TFIDF) \cite{rajaraman2011mining} comes under this category. Other filter methods that originate from information theory are, Mutual Information (MI) \cite{taira1999feature,tang2019feature}, Information Gain (IG) \cite{yang1997comparative}, CHI \cite{rogati}, ANOVA F measure \cite{elssied2014novel}, Bi-Normal Separation (BNS) \cite{forman2003extensive} and the GINI method \cite{shang2013feature}. They use hypothesis testing, contingency tables, mean and variance scores, conditional and posterior probabilities for selecting the features.

The wrapper method \cite{maldonado2009wrapper} use a search strategy to construct each possible subset, feeds each subset to the chosen classifier, and then evaluates the classifier's performance. These two steps are repeated until the desired quality of the feature subset is reached. The wrapper approach achieves better classification accuracy than filter methods. However, the time taken by the wrapper method is very high when compared to filter methods. 

Embedded methods complete the FS process within the construction of the machine learning algorithm itself. In other words, they perform feature selection during the model training. An embedded method is Decision Tree (DT) \cite{quinlan1986induction}. In DT, while constructing the classifier, DT selects the best features/attributes that may give the best discriminative power. 

Hybrid methods are robust and take less time when compared to the wrapper and embedded methods. They combine a filter method with a wrapper method during the feature selection process. The HYBRID model \cite{gunal2012hybrid} employs a combination of filter methods to select to rank the features and then a wrapper method to the obtained final features set. Our FS method is similar to the
HYBRID model. 

A detailed report on the benefits of using the feature selection methods in the pipeline with traditional classifiers is presented in \citet{deng2019feature,forman2003extensive}.

Apart from using traditional classification methods, deep feature selection using neural networks were also proposed. These models use deep neural network autoencoders for the feature set reduction and text generation \cite{mirzaei2019deep,han2018autoencoder}.

\citet{lam1999feature} studies the effect of feature set reduction before applying the neural network classifiers. The paper uses a multi-layer perceptron (MLP) classifier in combination with filter-based
FS method. \citet{alkhatib2017multi} proposes the use of neural network-based feature selection and text classification. Our work comes under this category.

\section{Proposed Pipeline}
\label{sect:method}
In this section, we present our feature selection and neural network pipeline.

The feature selection and neural network pipeline start with selecting a good tokenizer to tokenize the data and create a feature set. The tokenizer used for our feature selection is the Sentencepiece tokenizer  \cite{kudo2018sentencepiece}. Sentencepiece tokenizer implements subword units
by using  byte-pair-encoding (BPE) \cite{sennrich2015neural} and unigram language model \cite{kudo2018subword}. In the feature subset generation, we considered a hybrid feature selection method known as HYBRID \cite{gunal2012hybrid}. It has proved that a combination of the features selected by various methods is more effective and computationally faster than the features selected by individual filter and wrapper methods. Similar to the HYBRID model, we used three filters to obtain the relevancy score. The filters we considered were CHI2, ANOVA-F, and MI. These filters calculate the relevancy between the word and the class labels. 

Before feature selection, we used the Bag-of-Words(BoW) model to vectorize the data. In the BoW model, each feature vector is represented by $TF\_IDF$ scores. 

Then we used statistical measures such as $\chi^2$, ANOVA-F, and MI for obtaining feature scores. 

$\chi^{2}$ \footnote{A detailed explanation and a simple example of $\chi^{2}$ is given at https://www.mathsisfun.com/data/chi-square-test.html} is a statistic to measure a relationship between feature vectors and a label vector.

Analysis of Variance (ANOVA\footnote{A detailed explanation for ANOVA is given in https://towardsdatascience.com/anova-for-feature-selection-in-machine-learning-d9305e228476.}) is a statistical method used to check the means of two or more groups that are significantly different from each other. 

Mutual Information (MI\footnote{A simple explanation and working python example of MI is available at https://machinelearningmastery.com/information-gain-and-mutual-information/}) is frequently used to measure the mutual dependency between two variables. 

Using different statistical methods, we measured the relevance of each feature. We then aggregate the relevance scores of all satistical methods for each feature. The relevance of a feature $x_i$ is given by,

\begin{multline}
\label{relevance}
 Relevancy(x_{i}) =
 \\ \chi^2(x_{i}) + ANOVA(x_{i}) + MI(x_{i})
\end{multline}

\begin{figure*}[!h]
\centerline{\includegraphics[scale= 0.55]{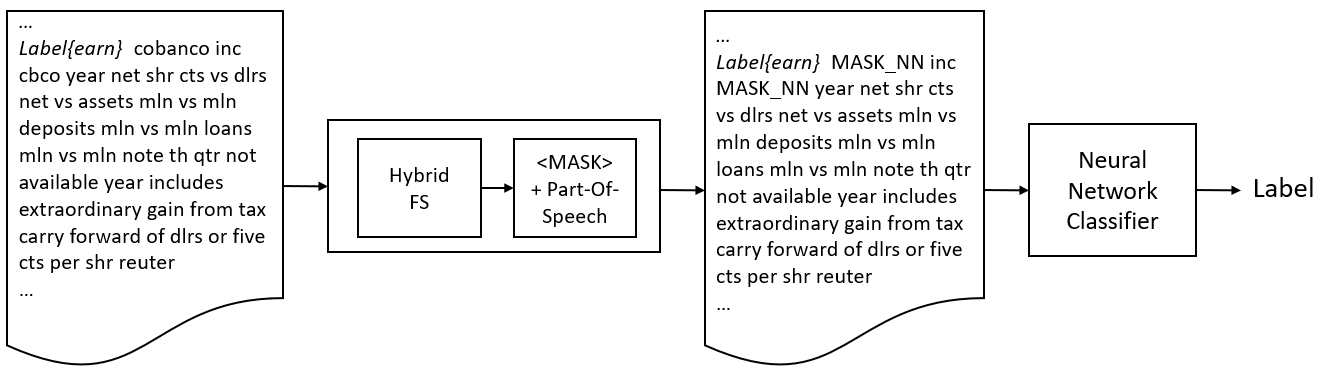}}
\caption{Modifying text by masking the low ranked words}
\label{fig:fsm2}
\end{figure*}

\begin{figure}[!h]
\centerline{\includegraphics[scale= 0.45]{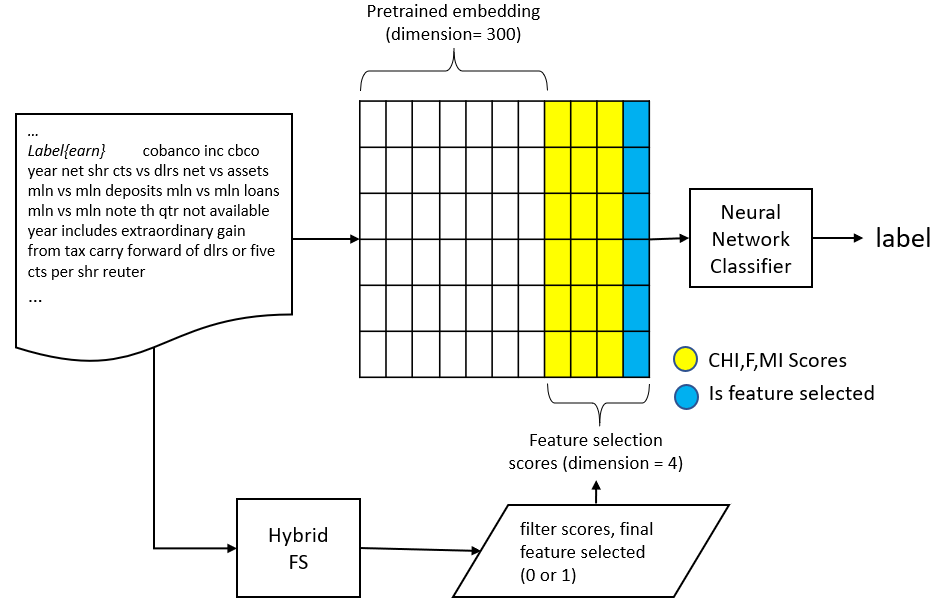}}
\caption{Meta-Embeddings, including feature scores along with word embeddings}
\label{fig:fsm3}
\end{figure}

Instead of an LR classifier given in the HYBRID model, we used the fastText classifier \cite{joulin2016bag} for the feature selection. We used the fastText classifier as it is often on par with deep learning classifiers in terms of accuracy and performs faster computations. The fastText classifier treats the average of word embeddings as document embeddings, then feeds document embeddings into a feed-forward NN or a multinomial LR classifier. We used pre-trained fastText word embeddings \cite{grave2018learning} while training a classifier.

To get the final features list, we sorted the normalized, aggregated value in descending order and divided the entire feature space into k sets. In our model, we divided the sorted feature space into 20 sets. The value of k is fixed to 20 using a trial and error basis.  We take the first set as the vocabulary of the classifier. We then trained the classifier and noted its accuracy. In the second iteration, we considered the vocabulary as the combination of first and second sets. Similarly, the third set has the vocabulary of the first three sets combined. We repeated the process until all the lists are exhausted. The set of features that resulted in a better classification metric is considered as the final feature set.

According to the proposed FS method, the final feature set is considered relevant, and they are necessary to perform the text classification. In contrast, the other features have little to no effect on the text classification or might degrade the classifier's performance. 

After feature subset generation, we propose three methods for including the feature selection information before training the neural network classifiers.

\begin{enumerate}

 \item \textbf{Method 1 (Selecting only the relevant features)\footnote{This method is already used while selecting the final features set by the fastText classifier.}:} Like traditional classification algorithms, we select only the relevant features that are estimated to be important by the feature selection method before training a neural network classifier. 
 
 \item \textbf{Method 2 (Masking the features that were given low importance by our FS method):} We felt that removing the features given low rank by our FS method might disturb the original data's grammatical structure, thus disturbing the word to word dependencies. We masked the low ranked words with the help of $<MASK>+POS(word)$ tag. $<MASK>$ word masks the low ranked word, and POS preserves the word's part of speech. The visual representation of method 2 is shown in figure \ref{fig:fsm2}
 
 \item \textbf{Method 3 (Meta Embeddings):} As shown in figure \ref{fig:fsm3}, we pass the relevancy and feature selection information along with embeddings in this method. Each slot holds the filter scores, i.e., CHI, ANOVA, MI scores of each feature. The last slot holds a 1 or 0 value. 1 is used for the selected features, and 0 is used for low ranked features that were not selected by our hybrid feature selection approach. 
\end{enumerate}

We analyzed and evaluated the above methods with various state-of-the-art NN classifiers on the benchmark datasets.

\subsection{Experiment}

In this section, we evaluated our feature selection and neural network pipeline on two tasks. We wanted to determine:
\begin{itemize}
 \item If the pipeline decreases the training time of the classifier
 \item If it helps in obtaining better local optima, thus improving the classification accuracy. 
\end{itemize}

We tested our pipeline across multiple state-of-the-art text classification algorithms.

\begin{enumerate}
 \item \textbf{CNN:} \cite{kim2014convolutional} This convolutional neural network-based text classifier is trained by considering pre-trained word vectors.
 \item \textbf{Bi-LSTM:} \cite{liu2016recurrent} A two-layer, bi-directional LSTM text classifier with pre-trained word embeddings as input was considered for the task of text classification.
 \item \textbf{fastText:} \cite{joulin2016bag} This is a simple, efficient, and the fastest text classification method. It treats the average of word/n-grams embeddings as document embeddings, then feeds document embeddings into a linear classifier.  
 \item \textbf{Text GCN:} \cite{yao2019graph} Builds a heterogeneous word document graph for a whole corpus and turns document classification into a node classification problem. It uses GCN \cite{kipf2017semi} to learn word and document embeddings.
 \item \textbf{DocBERT:} \cite{adhikari2019docbert} A fine-tuned BERT model for document classification. The BERT model \cite{devlin2018bert} uses a series of multiheaded attention and feedforward networks for various NLP tasks.
\end{enumerate}

\subsection{Datasets}
\label{ssec:datasets}

We ran our experiments on three widely used benchmark corpora and multilingual corpora. They are 20Newsgroups(20NG), R8, and R52 of Reuters 21578 and MLMRD.
\begin{itemize}
 \item The 20NG dataset contains 18,846 documents divided into 20 different categories. 11,314 documents were used for training, and 7,532 documents were used for testing. 
 \item R52 and R8 are two subsets of the Reuters 21578 dataset. R8 has 8 categories of the top eight document classes. It was split into 5,485 training and 2,189 test documents. R52 has 52 categories and was split into 6,532 training and 2,568 test documents. 
 \item MLMRD is a Multilingual Movie Review Dataset. It consists of the genre and synopsis of movies across multiple languages, namely Hindi, Telugu, Tamil, Malayalam, Korean, French, and Japanese. The data set is minimal and unbalanced. It has 9 classes and a total of 14,997 documents. The data was split into 10,493 training and 4,504 test documents.
\end{itemize}

We first preprocessed all the datasets by cleaning and tokenizing. The tokenizer used is the fastText tokenizer.

For baseline 1 models, we used multilingual fastText embeddings \cite{grave2018learning} of dimensionality 300, and baseline 2 models had the dimensionality of 304. We used default parameter settings as in their original papers for implementations. For calculating TFIDF, CHI2, ANOVA-F, MI scores, we used the scikit-learn library \cite{scikit-learn}. For POS tagging, we used the NLTK \cite{bird2009natural} pos tagger.

All the neural network models were run on the GPU processor on the Windows platform with NVIDIA RTX 2070 graphics card.

\section{Performance}

\begin{table}[!ht]
\centering
\begin{tabular}{|l|l|l|}
\hline
\textbf{Datasets}                             & Our FS               & HYBRID FS                  \\
\hline
20Newsgroups                  & \textbf{81.27\%}        & 77.34\%    \\
R8                            & \textbf{96.94\%}          & 93.79\%      \\
R52                           & \textbf{92.72\%}            & 86.43\%      \\
MLMRD                      & \textbf{47.09\%}           & 42.98\%      \\
\hline
\end{tabular}
\caption{The classification accuracy of our FS model when compared to the HYBRID model.}
\label{tab:fs_compare}
\end{table}

In our work, we modified the HYBRID \shortcite{gunal2012hybrid} feature selection model by changing the LR classifier to the fastText classifier. We selected the fastText classifier in the feature selection process because of its fast learning ability of a NN model compared to the traditional ML classifiers and other neural network classifiers \cite{joulin2016bag} without any decrease in classification accuracy. The neural network classifiers such as MLP, CNN, RNN, transformer, and GCN models achieve better classification accuracy when compared to traditional ML classifiers, but their training time is very high.

Using a fastText classifier during feature selection, we observed that our model performed better on all the benchmark datasets than the HYBRID model. The results are shown in table \ref{tab:fs_compare}. The fastText classifier's use helped the model obtain better relevant features, increasing the current feature selection model's accuracy compared to the HYBRID model. 

\begin{table*}[!ht]
\begin{tabular}{|l|l|l|l|l|}
\hline
\textbf{Datasets}       & \textbf{20Newsgroups} & \textbf{R8}      & \textbf{R52}     & \textbf{MLMRD}    \\
\hline
\textbf{Baseline 1 \& 2}     & 1,01,631 (V)                              & 19,956 (V)       & 26287 (V)        & 94073 (V)         \\
\textbf{Method 1} & 25732 (0.25V)                             & 17364 (0.87V)    & 22372 (0.85V)    & 52015 (0.55V)     \\
\textbf{Method 2} & 25732+30 (0.25V)                          & 17364+30 (0.87V) & 22372+30 (0.85V) & 52015+143 (0.55V) \\
\textbf{Method 3} & 1,01,631 (V)                              & 19,956 (V)       & 26287 (V)        & 94073 (V)         \\

\hline
\end{tabular}
\caption{The vocabulary size in all the FS inclusion methods when compared to the baselines. ``V'' is denoted as the vocabulary size of the actual data. Baselines 1,2, and method 3 have no change in vocabulary. However, using our FS method, the vocabulary is reduced to a maximum of 75\% (for 20Newsgroups data). Other datasets have seen a 13\% to 45\% decrease in vocabulary size. We can see an increase in vocabulary from method 1 to method 2. It is due to the additional vocabulary resulted from the mask words when they are accompanied by pos tags. Here Penn Treebank POS tagset is used.}
\label{tab:vocab}
\end{table*}

\begin{table*}[!h]
\centering
\begin{tabular}{|l|l|l|l|l|l|l|}
\hline
\textbf{Datasets} & \textbf{Method} & \multicolumn{5}{c|}{\textbf{Classifier(s)}}                                                                     \\ \hline
                  &                 & \textbf{CNN}     & \textbf{Bi-LSTM} & \textbf{fastText} & \textbf{DocBERT}                  & \textbf{Text GCN} \\ \hline
20Newsgroups      & Baseline 1      & 79.31\%          & 73.60\%          & 81.04\%           & \textbf{90.19\%}                  & 86.13\%           \\
                  & Baseline 2      & 79.46\%          & 74.25\%          & 82.44\%           & NA                                & 86.23\%           \\
                  & Method 1        & 78.27\%          & 73.44\%          & 81.27\%           & 89.37\%                           & \textbf{86.25\%}  \\
                  & Method 2        & 77.29\%          & 70.48\%          & 80.14\%           & 88.43\%                           & 85.65\%           \\
                  & Method 3        & \textbf{80.59\%} & \textbf{76.57\%} & \textbf{84.48\%}  & NA                                & 86.15\%           \\ \hline
R8                & Baseline 1      & 97.24\%          & 92.70\%          & 96.13\%           & \textbf{97.62\%} & 96.80\%           \\
                  & Baseline 2      & 97.37\%          & 93.82\%          & 96.50\%           & NA                                & \textbf{96.94\%}  \\
                  & Method 1        & \textbf{97.39\%} & 93.74\%          & 96.94\%           & 97.44\%                           & 96.28\%           \\
                  & Method 2        & 96.57\%          & 94.34\%          & 96.07\%           & 97.44\%                           & 96.85\%           \\
                  & Method 3        & \textbf{97.39\%} & \textbf{96.74\%} & \textbf{97.18\%}  & NA                                & \textbf{96.94\%}  \\ \hline
R52               & Baseline 1      & 94.78\%          & 87.53\%          & 92.02\%           & 92.95\%                           & 93.56\%           \\
                  & Baseline 2      & \textbf{94.84\%} & 90.79\%          & 92.76\%           & NA                                & 93.64\%           \\
                  & Method 1        & 94.29\%          & 87.47\%          & 92.72\%           & \textbf{93.10\%}                  & 92.97\%           \\
                  & Method 2        & 91.71\%          & \textbf{91.90\%} & 90.30\%           & 92.10\%                           & 93.19\%           \\
                  & Method 3        & \textbf{94.84\%} & 91.48\%          & \textbf{92.83\%}  & NA                                & \textbf{93.74\%}  \\ \hline
MLMRD             & Baseline  1     & 47.63\%          & 46.43\%          & 46.92\%           & \textbf{53.11\%}                  & 47.62\%           \\
                  & Baseline 2      & 47.79\%          & 47.43\%          & 48.92\%           & NA                                & 49.62\%           \\
                  & Method 1        & 44.98\%          & 44.82\%          & 47.09\%           & 51.90\%                           & 46.58\%           \\
                  & Method 2        & 44.63\%          & 44.05\%          & 46.61\%           & 50.90\%                           & 46.98\%           \\
                  & Method 3        & \textbf{48.44\%} & \textbf{49.13\%} & \textbf{49.55\%}  & NA                                & \textbf{51.50\%}  \\ \hline
\end{tabular}
\caption{Test accuracy on various neural network classifiers for the task of document classification. As the BERT model used is a fine-tuned one, we did not modify the model.}
\label{tab:result}
\end{table*}

As mentioned above, we used the training time-taken and test accuracy as the metrics to evaluate our approach. The accuracy and training time are recorded by running the model 10 times, and the average of the metrics was presented.

\subsection{Effects of our methods on classification accuracy}
Table \ref{tab:result} demonstrates the accuracy of feature selection methods on NN classifiers. 

When methods 1 and 2 were used, there is a slight decrease in classification accuracy because the first two methods lost semantic connection among words. Thus, the classification performance is degraded. Also, some words which were relevant to the classifier were masked out during the FS method. Whereas in method 3, including the feature selection scores with word-embeddings, has shown a significant improvement in accuracy on all the datasets.

Compared to the other datasets, the 20NG dataset has seen a significant decrease in vocabulary size. The vocabulary was decreased by 75\%. However, eliminating those features did not affect the accuracy of the classifier for methods 1 and 2.

Introducing the masked features in method 2 shown an increase in accuracy only in the Bi-LSTM method as this method considers word dependencies while training a classifier.

Including the feature selection scores along with the word-embeddings improved the classification accuracy on all the datasets. The feature selection metadata helped the neural network classifier learn a better relationship between the words and classes and improve the classifier's accuracy by reaching better local optima.

In R8 and R52 datasets, we have seen an increase in accuracy using method 1 because our hybrid FS method worked better on these datasets by removing the noisy words without disturbing the relevant words. The maximum improvement in accuracy is shown in the R8 dataset, with a +4\% increase in classification accuracy.

Our approach did not show any better results on MLMRD datasets as this dataset has a limited number of documents to train and test the data for some languages (Telugu, Tamil, Malayalam, Korean). Reducing vocabulary size by the FS method decreased the classification accuracy. 

\subsection{Effects of our methods on training time}

The pictorial representation of time taken by the classifiers for all the datasets is given in appendix B of the supplementary material.  

The time taken by method 1 is lower than in all baseline models. In method 1, as the text is modified by considering only relevant features, the vocabulary size is reduced, and the sentence length is reduced. It resulted in the more accelerated training of the neural network.

The time taken by baseline 2 and method 3 is similar because of the same embedding dimensionality of 304, but method 3 has achieved local optima a few epochs before compared to baseline 2, resulting in a time decrease of a few seconds. This phenomenon is attributed to the use of feature selection scores along with word embeddings.

Method 2 has shown an increase in training time even though the vocabulary is decreased because of 2 factors. 
\begin{enumerate}
\item The masking of features created unknown words in the data, and the classifier has to be trained to learn the representation of masked words, whereas the other words had pre-trained embeddings.
\item Apart from vocabulary, the neural network training time also depends on the input batch size given to the network and the length of the sentence in each batch. Because of the masked words, there is no decrease in either batch size or the sentence length. So the masking of data did not decrease the training time of the classifier.
\end{enumerate}

On the contrary, the Text GCN model has shown a decrease in training time because the classifier computes heterogeneous graph embeddings of each word based on the textual data before classification. It did not use any pre-trained embeddings.

In method 3, there is a slight increase in training time because of increased vocabulary size due to the inclusion of feature selection metadata. 

Of all the NN classifiers, the Text GCN model had shown a maximum decrease in training time by 488 sec when method 1 was used on 20NG data. As the Text GCN operates on building a graph on the complete vocabulary of data, the time taken by the method to build the graph is reduced significantly by reducing the vocabulary size. It is followed by the DocBERT and Bi-LSTM on 20NG data with a decrease in training time by 480 and 394 sec. Text GCN and Bi-LSTM have shown a significant decrease in training time on all the datasets. On the contrary, fastText and CNN are very fast while training the NN model. The training time of such models was unchanged when our method 1 was used.

When compared to all the classifiers, DocBERT achieved better results because of its evolutionary multi-headed attention and transformer models. As the Text GCN captures both local and word embeddings by constructing a heterogeneous graph, their results were better than those of the CNN and Bi-LSTM models, which work only on local word dependencies. As we increased the size of the embedding in FS method 2, this increased the dimensionality of vocabulary, resulting in the classifier's increased training time.

\section{Conclusion}

In our work, "Does a Hybrid NN FS Model Improve Text Classification?", we used the NN based hybrid FS method to extract relevant features and used NN classifiers for text classification. We extracted the relevant or high ranked features using filter-based methods and a fastText classifier. We then proposed three methods on how the feature selection can be included in the NN classification process. First, modifying the corpus by considering only relevant features. Second, modifying the data by masking the low ranked features, and the third method introduces feature selection information along with word embeddings. We observed that method 1 had shown a significant reduction in training time when large datasets or slower models are used, accompanied by a minimal change in classification accuracy. By introducing $MASK+P0S(word)$, we inferred that the masked word was a burden to the classifier, and it always tried to adjust the word embeddings, which resulted in increased epoch time during training and a slightly negative effect on classification accuracy. Whereas method 3 has shown no effect on decreasing the training time, it has shown a maximum of $ 4\%$  increase in the classification accuracy compared to baseline. It proved that introducing feature scores along with pre-trained word embeddings while training the NN classifier is beneficial.

Instead of opting for random naive vocabulary reduction techniques such as using min\_df and max\_df (minimum and maximum document frequency) for selecting features, by using FS methods, we can calculate the relevance of the word beforehand and use that as metadata to the NN classifier.  When the datasets are huge, these methods are of more significance. We can use the modified data while tuning the hyperparameters. Then we can use the real data to train and evaluate the model. Even in the critical domain datasets such as ``medical'', we cannot rely on removing a word based on min\_df and max\_df scores. Each word in those datasets should be treated with utmost significance. FS methods help in such scenarios by calculating the word's relevance and helps maintain better vocabulary before training neural network classifiers.

\bibliography{references}
\bibliographystyle{acl_natbib}

\end{document}


\appendix

\section{Discussion}
\label{sec:discussion}
As mentioned in the literature survey, there are feature selection models that use NN for selecting features. Such as variational and adversarial autoencoders. However, these methods take considerable time to learn the features that contradict our primary goal, i.e., reducing the classifier's training time. So we have resorted to using the fastest machine learning classifier ``fastText'' for selecting features.

\section{Time taken}
\label{sect:tt}
This section presents the pictorial representation of  training time taken by various classifiers on all the methods for each dataset. The pictorial representations start from the next page.

\begin{figure*}[!ht]
\centerline{\includegraphics[scale= 0.65]{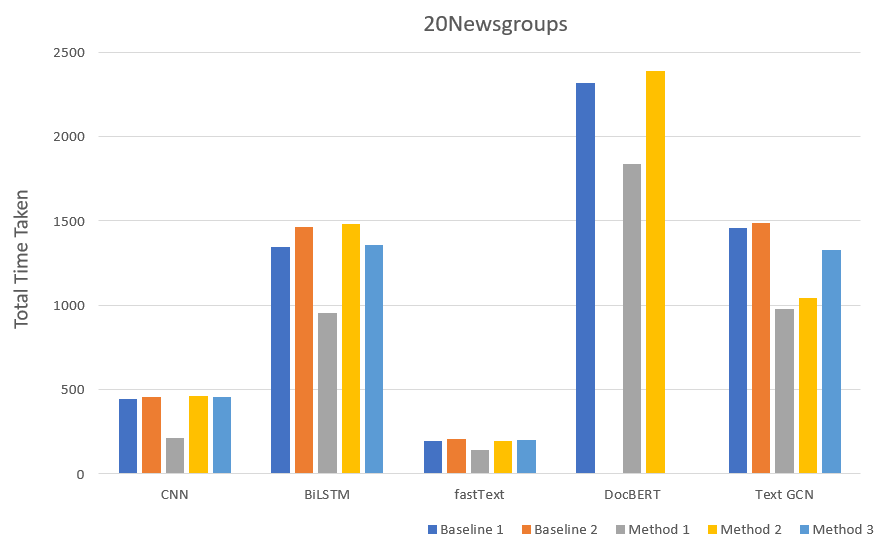}}
\caption{Time taken by various classifiers on 20Newsgroups dataset}
\label{fig:20ng_tt}
\end{figure*}

\begin{figure*}[!ht]
\centerline{\includegraphics[scale= 0.66]{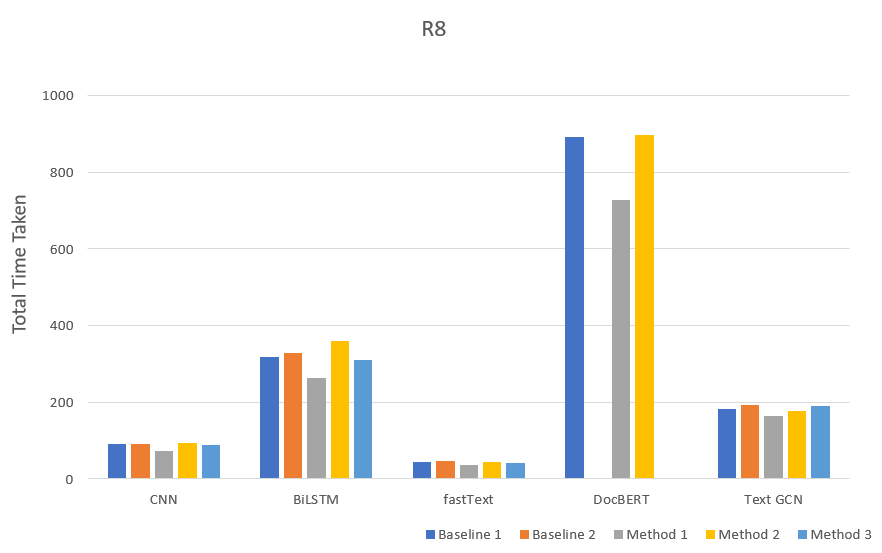}}
\caption{Time taken by various classifiers on R8 dataset}
\label{fig:r8_tt}
\end{figure*}

\begin{figure*}[!ht]
\centerline{\includegraphics[scale= 0.65]{r52_tt.png}}
\caption{Time taken by various classifiers on R52 dataset}
\label{fig:r52_tt}
\end{figure*}

\begin{figure*}[!ht]
\centerline{\includegraphics[scale= 0.66]{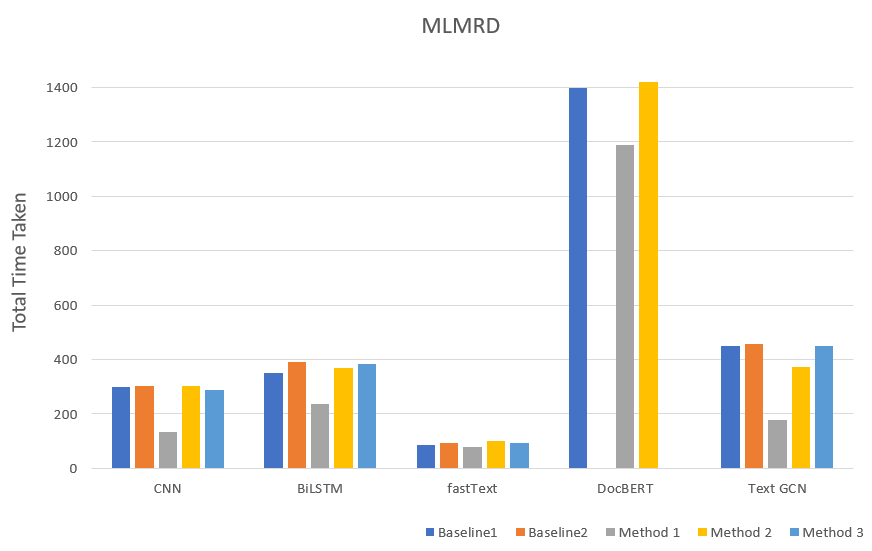}}
\caption{Time taken by various classifiers on MLMRD dataset}
\label{fig:mlmrd_tt}
\end{figure*}